\documentclass[conference]{IEEEtran}
\IEEEoverridecommandlockouts
\usepackage{cite}
\usepackage{amsmath,amssymb,amsfonts}
\usepackage{algorithmic}
\usepackage{graphicx}
\usepackage{textcomp}
\usepackage{float}
\usepackage{xcolor}
\usepackage{makecell}
\usepackage{multirow}
\usepackage{adjustbox}
\usepackage{hyperref}
\hypersetup{
    colorlinks=false,
    }

\def\BibTeX{{\rm B\kern-.05em{\sc i\kern-.025em b}\kern-.08em
    T\kern-.1667em\lower.7ex\hbox{E}\kern-.125emX}}

\makeatletter
\newcommand{\linebreakand}{%
  \end{@IEEEauthorhalign}
  \hfill\mbox{}\par
  \mbox{}\hfill\begin{@IEEEauthorhalign}
}
\makeatother

\begin{document}

\title{End-to-End Optical Character Recognition for Bengali Handwritten Words}


\author{
    \IEEEauthorblockN{Farisa Benta Safir\IEEEauthorrefmark{1}, Abu Quwsar Ohi\IEEEauthorrefmark{1}, M.F. Mridha\IEEEauthorrefmark{1}, Muhammad Mostafa Monowar\IEEEauthorrefmark{2}, Md. Abdul Hamid\IEEEauthorrefmark{2}}
    \IEEEauthorblockA{\IEEEauthorrefmark{1}\textit{Department of Computer Science \& Engineering}, \textit{Bangladesh University of Business \& Technology,}, \textit{Dhaka, Bangladesh}
    \\farisabentasafir@gmail.com, \{quwsarohi, firoz\}@bubt.edu.bd}
    \IEEEauthorblockA{\IEEEauthorrefmark{2}\textit{Department of Information Technology}, \textit{Faculty of Computing \& Information Technology}, \textit{King Abdulaziz University,}\\ \textit{Jeddah 21589, Kingdom of Saudi Arabia}
    \\\{mmonowar, mabdulhamid1\}@kau.edu.sa}
}

\maketitle

\begin{abstract}
    Optical character recognition (OCR) is a process of converting analogue documents into digital using document images. Currently, many commercial and non-commercial OCR systems exist for both handwritten and printed copies for different languages. Despite this, very few works are available in case of recognising Bengali words. Among them, most of the works focused on OCR of printed Bengali characters. This paper introduces an end-to-end OCR system \footnote{The code is available at \href{https://github.com/QuwsarOhi/bengali_word_ocr}{https://github.com/QuwsarOhi/bengali\_word\_ocr}} for Bengali language. The proposed architecture implements an end to end strategy that recognises handwritten Bengali words from handwritten word images. We experiment with popular convolutional neural network (CNN) architectures, including DenseNet, Xception, NASNet, and MobileNet to build the OCR architecture. Further, we experiment with two different recurrent neural networks (RNN) methods, LSTM and GRU. We evaluate the proposed architecture using BanglaWritting dataset, which is a peer-reviewed Bengali handwritten image dataset. The proposed method achieves $0.091$ character error rate and $0.273$ word error rate performed using DenseNet121 model with GRU recurrent layer.
\end{abstract}


\section{Introduction}
\label{introduction}
Technical advancement has led to converting analogue data into digital ones. Digital data has a wide range of automation support, such as searching text from documents, modification, context extraction, etc. Further, technical advancement with the help of artificial intelligence has enabled combating against pandemics \cite{alsamhi2020blockchain,alsamhi2021blockchain}, optimal mitigation, recognition of fabric elements \cite{ohi2021fabricnet}, intuitively optimal solutions \cite{ohi2020exploring} and so on. Hence, digitalising paper documents would benefit more extensive usage of information.

OCR scans an image and generates a machine-encoded file. OCR is a popular technology used in automated data capture solutions and document classifications. There are different types of OCR systems, including intelligent word recognition, intelligent character recognition, optical word recognition, optical character recognition, and optical mark recognition. It has a broad area of research due to its usefulness. OCR systems are used for converting books, documents, and images into a computerised file. OCR is used in diverse domains including banks, post-offices, defence organisations, license plate recognition, reading aid for the blind, library automation, language processing, multimedia system design, education institute, etc. People can use an OCR to reduce the complexity of digitising documents manually. OCR systems can benefit organisations by facilitating with better process speed, enhanced workforce, and lower costs. With the development of digital computers and scanning devices, OCR technology is improved in the middle of the 1950s. The OCR system has two major categories: typewritten and handwritten scripts. Typewritten scripts typed in computers before the process of recognition starts. On the other hand, handwritten scripts are written by humans and then recognised by the OCR system. Typewritten OCR systems are more likely easy to implement than handwritten OCR systems. Also, typewritten recognition systems' success rate is more than the handwritten ones, as they are less complicated and less variation is observed. Implementing an OCR system is not easy as machines cannot perceive information from an image like the human brain. Therefore, many researchers have made efforts to transform a document into a machine-readable file since the middle 1950s.

There are many established languages worldwide, like English, Chinese, Arabic, Japanese, Bengali, etc. Bengali is the fifth most-spoken native language and the seventh most spoken language by the world’s total number of speakers\cite{enwiki:1007067782}. Many different kinds of documents that include letters, textbooks, novels, official documents, legacy documents, newspapers, magazines, data entry forms, etc., are in Bengali that needed digitalisation. Many renowned Bengali handwritten pieces of literature also need to be computerised and stored. 

Bengali language has much more difficult characters because of its shapes than any other language.  In the Bengali language, it consists of 11 vowels (Shoroborno), 39 consonants (Benjonborno) and two or more characters combine to form new characters called compound (Juktoborno) characters. Thereby, building an OCR that recognises Bengali characters is more complicated than recognising any other language characters. Hence, many researchers have been working to develop an OCR system for identifying Bengali characters since the middle 1980s. Since then, Bengali OCR is a large field of research, and many researchers have proposed state-of-the-art solutions. But yet, to the best of our knowledge, there is no OCR that recognises handwritten Bengali words from images.

In this paper,  we propose a model that recognises handwritten Bengali words from images. Our approach incorporates end-to-end architecture that depends on CTC loss function. From an architectural perspective, we combine deep convolutional neural network, that works as a feature extractor from handwritten images. Further, the extracted features (scanned from left to right of a word image) is passed to a recurrent layer, specifically, LSTM, or GRU. The recurrent layers extract dependent features based on the previously seen image slices and produce high dimensional features. Finally, a fully connected layer generates a probability distribution of the final prediction.

The overall contribution of the paper includes:
\begin{itemize}
    \item We introduce an end-to-end word recognition system for the Bengali language. This is the first research endeavour that explores end-to-end strategy in Bengali OCR to the best of our knowledge.
    \item We investigate end-to-end strategy's performance with established Bengali dataset, BanglaWriting \cite{mridha2021banglawriting}. 
    \item As a feature extractor, we use four different baselines (Xception, NASNet, MobileNet, and DenseNet) and conclude that deeper architecture with residuals performs better in case of Bengali handwritten OCR.
\end{itemize}

The rest of the paper is constructed as follows: Section \ref{sec:related_work} highlights the works conducted in the domain of optical character recognition, specifically for the Bengali language. Section \ref{sec:methodology} explicates the proposed architecture and means undertaken to build the system. Section \ref{sec:experimental_setup} describes the overall experiments undertaken to evaluate the proposed method. Finally, Section \ref{sec:conclusion} concludes the paper.

\section{Related Work}
\label{sec:related_work}

Many complete OCR systems exist for Bengali scripts as it is a research topic since the 1980s. Different researchers have already done many remarkable jobs in this field. Some of these works are worth mentioning. B.B. Chowdhuri and U. Pal proposed “OCR in Bangla: an Indo-Bangladeshi language” \cite{pal1994ocr} and also suggested a complete printed Bengali OCR system \cite{chaudhuri1998complete} including the feature extraction process for recognition. 

J. U. Mahmud, et.al., proposed another complete OCR that recognises isolated and continuous printed multi-font Bengali characters \cite{mahmud2003complete}, achieving 98\% recognition rate in isolated characters and 96\% recognition rate in continuous characters.

A. Chowdury, et.al., proposed a better approach for Optical Character Recognition of Bengali Characters using neural networks \cite{chowdhury2002optical}. They also describe the efficient ways of involving line and word detection, zoning, character separations, and character recognition.

Hasnat M.A. et al. proposed a domain-specific OCR for Bengali script \cite{hasnat2008high}, using Hidden Markov Model (HMM) for character classification and added a particular error correcting module to handle the errors that occurred at the preprocessing level. Finally, when the word is formatted, they added a dictionary and defined rules to correct the probable errors.

All the aforementioned efforts have a significant limitation, i.e., they work with printed or typewritten scripts only, and not capable of recognising handwritten scripts. To overcome this limitation, many researchers have worked with Bengali handwritten scripts.

Pramanik R, Bag S proposed a novel shape decomposition-based segmentation technique to decompose the compound characters into prominent shape components \cite{pramanik2018shape}. They have claimed that this shape decomposition technique reduces the classification complexity in terms of less number of classes to recognise, and at the same time improves the recognition accuracy. Further, they used a chain code histogram feature set with a multi-layer perceptron (MLP) based classifier with backpropagation learning for classification.

M. Al Rabbani Alif, et.al. proposed a modified ResNet-18 architecture to recognise Bengali handwritten characters \cite{alif2017isolated}. They restructured the ResNet-18 architecture by adding dropout layers that boost the classification performance. They have applied their architecture on the BanglaLekha-Isolated dataset \cite{biswas2017banglalekha} and CMATERdb dataset \cite{das2014benchmark} and obtained an accuracy of $95.10$\% and $95.99$\%, respectively.

Rakshit P., et.al proposed a scheme for tri-level segmentation (line, word, and character) for Bengali handwritten scripts \cite{rakshit2018line}. They have achieved an average of $90.46$\% accuracy on line segmentation, $90.06$\% accuracy on word segmentation, and $50.55$\% accuracy on character segmentation for the dataset of $50$ Bengali handwritten text documents.

Hasan F., Shuvo S.N. et al. proposed a new methodology to recognise the character from continuous Bengali handwritten characters using CNN \cite{hasan2020bangla}. They take continuous Bengali handwritten text images as an input and then segment the input texts into their constituent words and segment each word into individual characters. They have used the EkushNet dataset model \cite{rabby2018ekushnet} which includes 50 basic characters, 10 character modifiers, 52 frequently used conjunct characters, 10 digits, and able to segment $95$\% words from text and $90$\% characters from the words.

There are few studies in the literature that only segments Bengali handwriting words. S. Basu, et al. \cite{basu2007fuzzy}, proposed a fuzzy technique for segmentation of handwritten Bengali word images. First, they identify the Matra(i.e., the longest straight-line connects multiple characters to make a Bengali word) using a fuzzy feature from the target word image. Then some parts of the Matra are identified as a segment point by using three fuzzy features. They use only 210 samples of handwritten Bengali words to run the experiment, and they claim $95.32$\% average accuracy.

Pramanik R., Bag S. \cite{pramanik2020segmentation}, proposed a method for recognising handwritten Bengali and Devanagari words that detect and correct skew present in words and then estimated the headline, segmenting the words into a meaningful pseudo character. This is the only research work that recognizes Bengali word from images as far as our knowledge. They extract three different statistical features and combine them and apply CNN-based transfer learning architecture. After that, they combine the identified pseudo characters to make the full word. They claimed $94.01$\% accuracy in recognising Bengali words from images in their proposed segmentation methodology. To run the experiment they have used 2000 Bengali word images from Cmaterdb dataset version 1.1.1 \cite{sarkar2012cmaterdb1} and 1.5.1 \cite{singh2018benchmark}, ICDAR 2013 Segmentation dataset \cite{stamatopoulos2013icdar}, and PHDIndic 11 dataset \cite{obaidullah2018phdindic_11}. However, 2000 data is insufficient to build and benchmark an OCR system for handwritten Bengali words. 

All of these proposed architectures depend on outdated strategies for Bengali OCR. Hence, this paper proposed an end-to-end approach that recognises handwritten Bengali words from handwritten word images. Further, experiments conducted in similar research work \cite{pramanik2020segmentation} use fewer words (at most 2000 words) in the evaluation phase. In contrast, we use 16975 words from BanglaWritting dataset to run the experiment. To build the system, we use different CNN and RNN architectures. Finally, we show a benchmark with the evaluated results.

\section{Methodology}
\label{sec:methodology}
\begin{figure}[H]
    \centering
    \includegraphics[width=\linewidth]{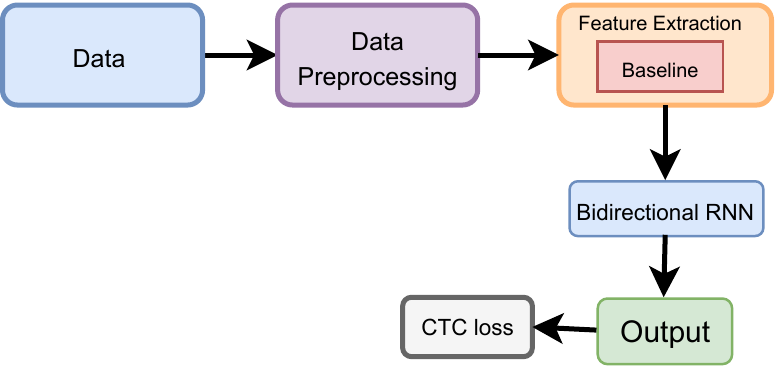}
    \caption{The block diagram illustrates the general workflow of the proposed methodology.}
    \label{fig:workflow}
\end{figure}

This method recognises handwritten Bengali words from word-level images. First, we extract features from the images, and then we use loss function to train the model and calculate the loss and error. The complete methodology is divided into the following steps: (a) Data collections and preprocessing, (b) Features extraction, and (c) Loss and error calculation. Fig \ref{fig:workflow}. represents a visual of the mentioned workflow. 

\subsection{Data preprocessing}
\label{sec:data_preprocessing}

\subsubsection{Reshape and normalization}
\label{sec:reshape_and_normalization}

Before fitting word-level image data into the architecture, we preprocessed the data to fulfil some conditions. First, each word-level image is reshaped into 50 by 200 pixels. Second, we consider the maximum number of characters in a word to be 10. Hence the word with characters of more than 10 is ignored from both text and image database. The image data are then normalized in the range of [-1, 1], as data normalization certifies a parallel data distribution in every input parameter. Each input image is normalized as,
\begin{equation}
    Normalize(D) = 
    \left( \begin{bmatrix} 
        d_{11} & \dots & d_{1m} \\
        \vdots & \ddots & \vdots \\
        d_{n1} & \dots & d_{nm} 
    \end{bmatrix} / 127.5 \right) - 1    
\end{equation}

Here $D$ is the single-channel word image matrix, $n$ is the number of rows, and $m$ is the number of columns of the word image matrix.

\begin{figure}[h]
    \centering
    \includegraphics[width=\linewidth]{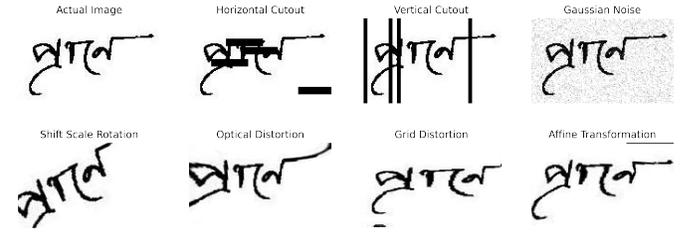}
    \caption{The picture illustrates the actual handwritten word-level image, and seven different augmentation policies explained in Section \ref{sec:augmentation}.}
    \label{fig:augmentation}
\end{figure}

\subsubsection{Augmentation}
\label{sec:augmentation}

Data augmentation is a popular technique to expand the variety of trainable data and to avoid overfitting while training models. Therefore we will use some augmentation techniques to create a diverse dataset. We use the open-source albumentation \cite{buslaev2020albumentations} library that implements various image augmentation strategies. Fig. \ref{fig:augmentation} represents the augmentation for one image. The implemented augmentation strategies are described below.

\begin{itemize}
    \item \textbf{Horizontal cutout} erases some part of the image by adding some random horizontally black box. 
    \item \textbf{Vertical cutout} erases some part of the image by adding some random vertical black box.
    \item \textbf{Gaussian noise} adds some dot randomly to create noise in the image.
    \item \textbf{Shift scale rotation} shifts and rotate the image within a given scale.
    \item \textbf{Optical distortion} distorts pixel patterns of handwritten images.
    \item \textbf{Grid distortion} elastically distorts written patterns, causing written lines to shrink and bend.
    \item \textbf{The Affine transformation} adds a regular grid of points on the image and randomly moves the neighbourhood of these points around.
\end{itemize}

\begin{figure}[h]
    \centering
    \includegraphics[width=\linewidth]{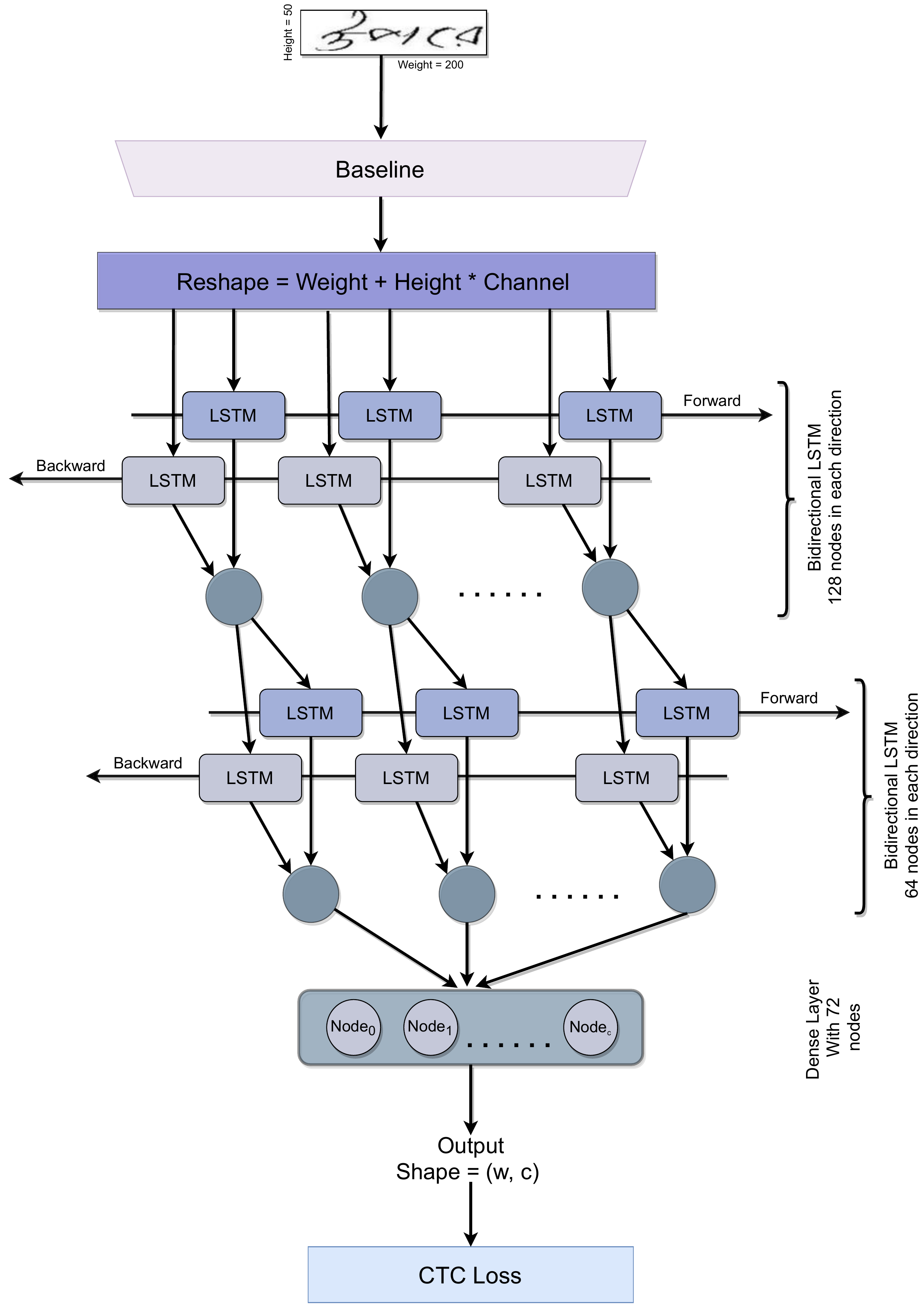}
    \caption{The picture exposes the neural network architecture of the developed end-to-end Bengali OCR system. The system contains one deep convolutional neural network as a baseline, followed by two bidirectional RNN layers.}
    \label{fig:model}
\end{figure}

\subsection{Features Extraction}
\label{sec:feature_extraction}
Feature extraction is the process of finding necessary information from an image to identify the word. The feature extraction process is segmented into two parts in our architecture: a baseline model and a stack of bidirectional RNN model. The input layer passes the image in the shape of $200\times50\times3$ to get the best results in our architecture. Then, the baseline model directly receives input images and extracts high dimensional features. To create the baseline model, we use famous CNN architectures rather than building our own to minimise the workload and get the full feature. We have modified the pre-trained convolutional model by deducting some layers, guaranteeing 25 left-to-right frames for the end-to-end method. Convolutional layers take input images, extract useful features and give a lower-dimensional output. To build the end to end system, we fix operational output as $25\times72$ (72 is the number of unique characters). Therefore, we experiment with four different convolutional architectures and evaluate their performance in the results analysis section. The features obtained from the baseline model is passed to the bidirectional RNN model. In the bidirectional RNN model, two bidirectional RNN layers are assigned to predict the word from the image by analysing the features’ sequence. Since general RNN has vanishing gradient problems, we use two modified RNN,  Long Short Term Memory(LSTM) \cite{hochreiter1997long} and Gated Recurrent Unit (GRU) to avoid vanishing gradient problems. We experiment with these two modified versions of RNN and evaluate their accuracy in the results analysis section. The bidirectional RNN layer’s output passes through a fully connected dense layer with a softmax activation function and provides the final output. The final output shape must be $25\times72$ to fit in the CTC loss function. Fig. \ref{fig:model} illustrates a block diagram of full architecture. 

\subsection{Loss and error calculation}
\label{sec:loss_and_error_calculation}

We used the Connectionist Temporal Classification (CTC) \cite{graves2006connectionist} loss function to train our model and calculate the loss and error throughout the training. The CTC loss function does not need aligned data. Instead, it works by adding all possible probability of alignments. As our handwritten data is not aligned, we use the CTC loss function. Also, CTC can train the whole model on its own. Therefore, we used the CTC loss function for our end to end architecture. 


\section{Experimental analysis}
\label{sec:experimental analysis}

\subsection{The Bengali handwritten dataset}
\label{sec:dataset}
\begin{figure}[h]
    \centering
    \includegraphics[width=\linewidth]{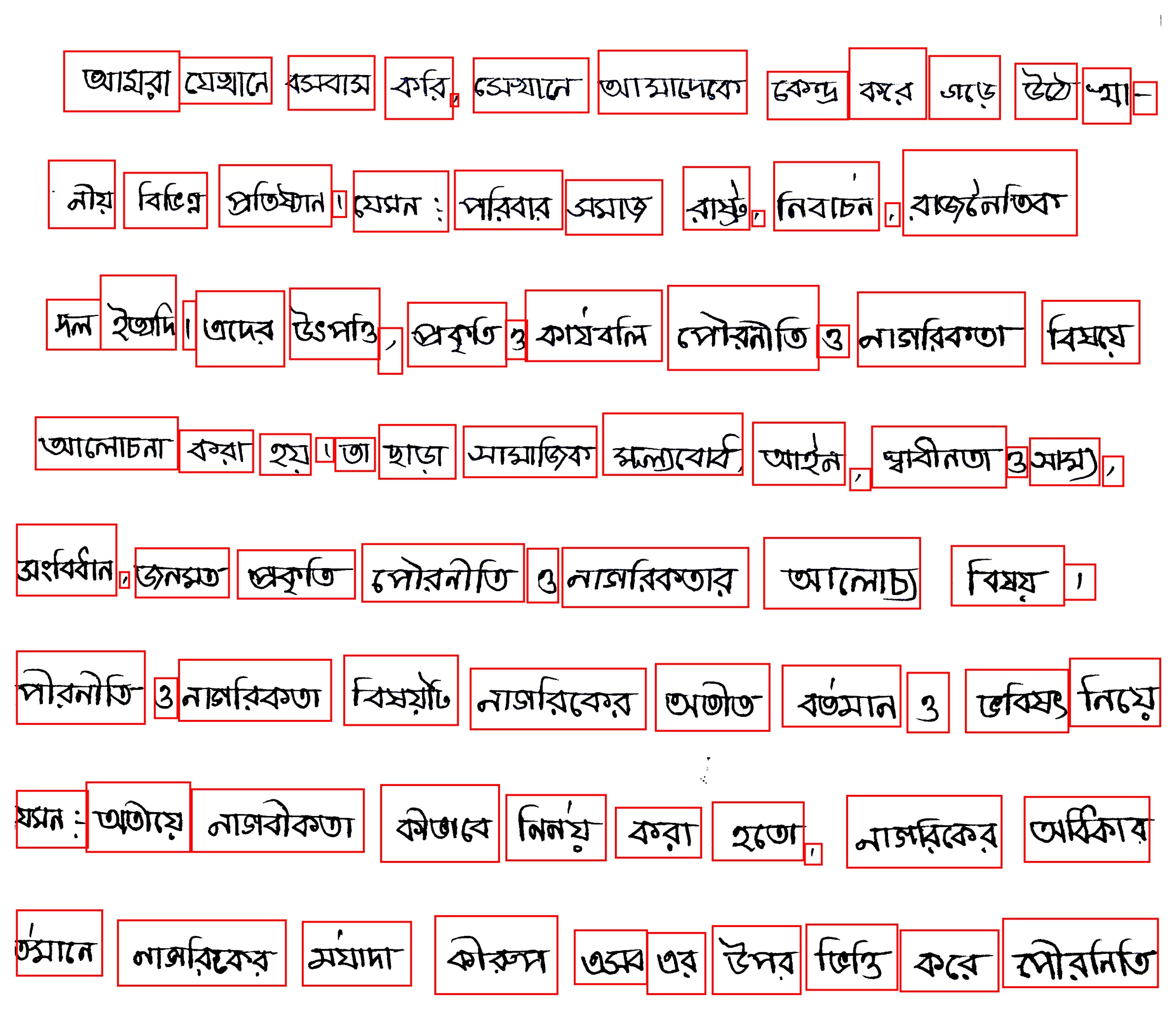}
    \caption{The image illustrates the word level data extracted from a single page handwriting.}
    \label{fig:dataset}
\end{figure}

To evaluate the proposed architecture, BanglaWritting \cite{mridha2021banglawriting} dataset is used. The dataset contains single-page handwriting of 260 different peoples. Every page includes a unicode to represent the writing and bounding box to bound each word. The dataset contains 21,234 words. However, we took 16975 words to run the experiment, resulting in 73 unique Bengali characters. Fig. \ref{fig:dataset} includes the example of images.

\subsection{Experimental setup}
\label{sec:experimental_setup}
The proposed architecture is implemented using Tensorflow \cite{abadi2016tensorflow}, Keras \cite{chollet2015keras}, Matplotlib \cite{hunter2007matplotlib},  NumPy \cite{van2011numpy} and Python \cite{van1991python}. Albumentation \cite{buslaev2020albumentations} is used to augment the dataset. CTC loss function is used to measure the loss and error of the architecture. The dataset is divided into train, validation, and test subsets, where each one contains $70$\%, $15$\%, and $15$\% of data, respectively. Each architecture is trained using a batch size of 16 with a maximum epoch limit of 1000 with 0.001 learning rate. Adam optimiser is used to train the full architecture.


\begin{table*}[!h]
    \centering
    \caption{The table illustrates the benchmarks implemented using different deep CNN architectures along with different RNN categories. FLOPs illustrate the architectures' computational complexity, whereas the CER and WER explain the character and word error rate, respectively.}
    \label{tab:benchmark}
    \adjustbox{width=\linewidth}{
    \begin{tabular}{|c|c|c|c|c|c|c|c|c|c|c|c|}
        \hline
        \multicolumn{2}{|c|}{\textbf{Baseline}} & \multirow{2}{*}{\textbf{\makecell{FLOPs\\(million)}}} &
        \multicolumn{3}{c|}{\textbf{Train}} &
        \multicolumn{3}{c|}{\textbf{Validation}} & \multicolumn{3}{c|}{\textbf{Test}} \\ \cline{1-2} \cline{4-12}
        
        \textbf{Model} & \textbf{RNN} && \textbf{Loss} & \textbf{CER} & \textbf{WER} & \textbf{Loss} & \textbf{CER} & \textbf{WER} & \textbf{Loss} & \textbf{CER} & \textbf{WER} \\ \hline \hline
        
        \multirow{2}{*}{Xception \cite{chollet2017xception}} & LSTM & 7.2 & 10.282 & 0.389 & 0.711 & 10.025 & 0.461 & 0.764 & 10.106 & 0.467 & 0.764 \\ \cline{2-12}
        & GRU & 7.4 & 12.891 & 0.543 & 0.853 & 11.946 & 0.565 & 0.857 & 12.060 & 0.565 & 0.858 \\ \hline
        
        \multirow{2}{*}{MobileNet \cite{howard2017mobilenets}} & LSTM & 2.8 & 2.808 & 0.002 & 0.012 & 3.948 & 0.107 & 0.285 & 3.847 & 0.107 & 0.291 \\ \cline{2-12}
        
        & GRU & 3.8 & 3.312 & 0.009 & 0.039 & 5.095 & 0.175 & 0.384 & 5.165 & 0.182 & 0.414 \\ \hline
        
        \multirow{2}{*}{\textbf{DenseNet121} \cite{huang2017densely}} & LSTM & 5.4 & 3.277 & 0.019 & 0.079 & 4.54 & 0.144 & 0.336 & 4.57 & 0.146 & 0.346 \\ \cline{2-12}
        
        & \textbf{GRU} & \textbf{6.4} & \textbf{2.806} & \textbf{0.010} & \textbf{0.048} & \textbf{3.49} & \textbf{0.091} & \textbf{0.260} & \textbf{3.463} & \textbf{0.091} & \textbf{0.273} \\ \hline
        
        \multirow{2}{*}{NASNetMobile \cite{zoph2018learning}} & LSTM & 1.9 & 3.870 & 0.039 & 0.141 & 5.176 & 0.177 & 0.409 & 5.319 & 0.189 & 0.425 \\ \cline{2-12}
        
        & GRU & 2.4 & 3.979 & 0.045 & 0.171 & 4.628 & 0.158 & 0.401 & 4.601 & 0.160 & 0.403 \\ \hline
    
    \end{tabular}}
\end{table*}

\subsection{Evaluation metrics}
\label{evaluation metrics}
Two evaluation metrics have been used to compare and evaluate the performance of our architecture. These are presenting as follows:

\begin{itemize}

\item \textbf{Character Error Rate:} Character Error Rate (CER) indicates the number of erroneous predictions made by the OCR system. To calculate the CER, we use the edit distance algorithm, which is used as follows,

\begin{equation}
    CER = \frac{S + I + D}{\text{Number of characters}}
\end{equation}

Here, $S$ means substitutions; $I$ means insertions, $D$ means deletions.

Substitutions occur when a character gets replaced within a word. Insertions happen when an extra character gets added in a word that was not in the actual word. Deletions is when a character gets removed from the word that was present in the actual word. 

\item \textbf{Word Error Rate:} Word Error Rate(WER) indicates the number of handwritten words that OCR does not recognize properly. The WER is calculated by comparing predicted words with testing data formalized as,
\begin{equation}
    WER = \frac{\text{Incorrect word predictions}}{\text{Total word predictions}}    
\end{equation}

\item \textbf{FLOPs:} Floating point operation per second used to measure the performance of the model. It calculates the number of arithmetic operations needed to run the deep learning model. Lower FLOPs indicate lower time complexity of the model.

\end{itemize}

\subsection{Results analysis}
\label{sec:result_analysis}
To build the end-to-end OCR system, we experiment with four famous architectures. Yet, some modifications are made in the architecture to sync the architectures with the current implementation properly. The modifications are discussed below:

\subsubsection{Baseline}
\label{sec:baseline}
Famous deep CNN architectures are used as the baseline of the model to extract useful information from images. In this paper, we focus on end-to-end architecture more than the baseline model. Hence, we used four different pre-trained convolutional models trained on ImageNet dataset that includes MobileNet \cite{howard2017mobilenets}, DenseNet121 \cite{huang2017densely}, Xception \cite{chollet2017xception}, and NASNetMobile \cite{zoph2018learning}. For the implementation, we used the Keras framework. We have deducted some layers of these architectures to reach a fixed operational output. Figure \ref{fig:mobileNet}, \ref{fig:densenet121}, and \ref{fig:xception} illustrates the modification of the MobileNet, DenseNet121, and Xception architecture, respectively. Furthr, the deductions are discussed below.

\begin{itemize}
    \item \textbf{MobileNet:} MobileNet is a famous convolutional architecture pre-trained on ImageNet dataset. It has a total of 29 deep convolutional layers. We use the first 11 deep convolutional layers from MobileNet.
    
    \begin{figure}[H]
    \centering
    \includegraphics[width=0.7\linewidth]{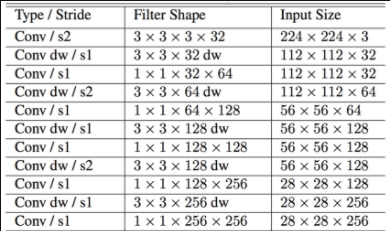}
    \caption{Layers of MobileNet architecture used as a baseline.}
    \label{fig:mobileNet}
\end{figure}

    \item \textbf{DenseNet121:} Densenet121 is a convolution architecture in which each layer is connected with the deeper layers. From Densenet121, We use the layers till dense block 2.
    
    \begin{figure}[H]
    \centering
    \includegraphics[width=0.7\linewidth]{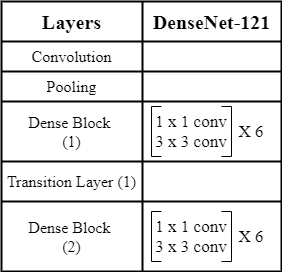}
    \caption{Layers of DenseNet121 architecture used as a baseline.}
    \label{fig:densenet121}
\end{figure}

    \item \textbf{Xception:} Xception is another convolution architecture. It has three flows; Entry flow, Middle flow and Exit flow. In our architecture, we use only the Entry flow without the last max-pooling layer. 
    
    \begin{figure}[H]
    \centering
    \includegraphics[width=0.7\linewidth]{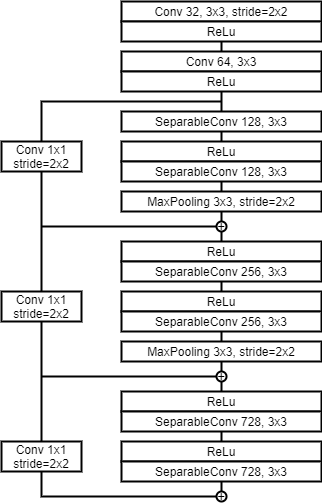}
    \caption{Layers of Xception architecture used as a baseline.}
    \label{fig:xception}
\end{figure}

    \item \textbf{NASNetMobile:} NASNet mobile is a convolutional architecture that can classify an image into 1000 categories. We use 71 activation layers, (defined as `activation\_71` in Keras framework) from NASNet mobile in our baseline. 
\end{itemize}

All of these, baseline architecture compared with each other in Bengali handwritten OCR, are presented in Table \ref{tab:benchmark}. The comparison represents the FLOPs, Loss, CER, WER of all the architectures calculated on the train, validation, and test dataset. The comparison shows that the end-to-end architecture achieves the best result using DenseNet121 with GRU recurrent layers. In DenseNet121, each dense convolutional layer is connected with a skip connection with the previous and later dense convolution layer. As the restructured network of DenseNet121 was substantially deep, the architecture performs according to the actual implementation. Hence, DenseNet121 achieves excellent performance. On the contrary, in case of Xception, most layers have been removed, resulting in heavy performance degradation. MobileNet and NASNet mobile perform a good result considering the CER. However, the WER values for both of the baselines are high. As both of the architectures include lesser residuals, they tend to lose some common character appearances. Therefore, although the models' CER values are low, the repetition of similar character errors result in higher WER. Hence, from the overall observation, DenseNet121 achieves a considerably better CER and WER than other architectures.

\section{Conclusion}
\label{sec:conclusion}
The paper presents an end to end OCR system that recognizes Bengali handwritten words from images. The proposed OCR system is implemented based on an end-to-end architecture and experimented with using a rich Bengali handwriting dataset called BanglaWritting. The baseline architecture is implemented and evaluated by four different pre-trained CNN architectures (i.e. MobileNet, Xception, DenseNet121, NASNet mobile). Further, we use two different bidirectional RNN (LSTM, GRU) types. DenseNet121 with GRU achieves the best results CER $0.09$ and WER $0.273$. However, to improve this system in the future, a better investigation neural network is required rather than a pre-trained network as a baseline model. Automatic word segmentation is also essential to build an end-to-end OCR system. The research work is a leap towards achieving more robust investigation and implementation related to Bengali OCR systems.

\bibliographystyle{unsrt}
\bibliography{reference}

\end{document}